\documentclass[letterpaper, 10 pt, conference]{ieeeconf}  

\IEEEoverridecommandlockouts                              
\usepackage{amsmath}    			
\usepackage{amssymb}
\usepackage{amsfonts}

\usepackage{graphicx}  				

\usepackage[croatian]{babel}  
\usepackage[utf8]{inputenc}  	
\usepackage[T1]{fontenc}
\usepackage{ae,aecompl}     	

\usepackage{microtype}				

\usepackage{subcaption}

\usepackage{tabularx}
\usepackage{booktabs}
\usepackage{enumerate}				
\usepackage{todonotes}				
\usepackage{dirtree}					
\usepackage{url} 

\usepackage{lmodern}
\usepackage{nccmath}
\usepackage[keeplastbox,nospread]{flushend}
\usepackage{scalerel,stackengine}
\stackMath
\newcommand\reallywidehat[1]{%
	\savestack{\tmpbox}{\stretchto{%
			\scaleto{%
				\scalerel*[\widthof{\ensuremath{#1}}]{\kern.1pt\mathchar"0362\kern.1pt}%
				{\rule{0ex}{\textheight}}
			}{\textheight}%
		}{2.4ex}}%
	\stackon[-6.9pt]{#1}{\tmpbox}%
}
\usepackage{mathptmx}
\usepackage{graphicx, subcaption}
\usepackage{array}
\usepackage{booktabs}
\usepackage[ruled, vlined, linesnumbered]{algorithm2e}
\usepackage{cleveref}
\usepackage{algpseudocode}
\usepackage{multirow, makecell, caption, hhline}

\usepackage{algorithm2e}
\usepackage[bottom]{footmisc}
\SetKwInput{KwPersistent}{Persistent variables}
\SetKwInput{KwLocal}{Local variable initializations}
\SetArgSty{textnormal}
\makeatletter
\newcommand{\nosemic}{\renewcommand{\@endalgocfline}{\relax}}
\newcommand{\dosemic}{\renewcommand{\@endalgocfline}{\algocf@endline}}

\makeatother

\makeatletter
\newcommand{\removelatexerror}{\let\@latex@error\@gobble}
\makeatother

\usepackage{calrsfs}
\DeclareMathAlphabet{\pazocal}{OMS}{zplm}{m}{n}
\DeclareMathOperator*{\argmax}{arg\,max}

\graphicspath{{./figures/}}
\usepackage{float}

\title{\LARGE \bf
A Multi-Resolution Frontier-Based Planner for Autonomous 3D Exploration 
}

\author{Ana Batinović, Tamara Petrović, Antun Ivanovic, Frano Petric, Stjepan Bogdan
	\thanks{Authors are with the University of Zagreb, Faculty of Electrical Engineering  and Computing, LARICS Laboratory for Robotics and Intelligent Control Systems, Unska 3, 10000 Zagreb, Croatia; {\tt\small ana.batinovic@fer.hr}}}


\DeclareUnicodeCharacter{2212}{-}
\begin{document}
\maketitle
\thispagestyle{empty}
\pagestyle{empty}

\begin{abstract}

In this paper we propose a planner for 3D exploration that is suitable for applications using state-of-the-art 3D sensors such as lidars, which produce large point clouds with each scan. The planner is based on the detection of a frontier - a boundary between the explored and unknown part of the environment - and consists of the algorithm for detecting frontier points, followed by clustering of frontier points and selecting the best frontier point to be explored. Compared to existing frontier-based approaches, the planner is more scalable, i.e. it requires less time for the same data set size while ensuring similar exploration time. Performance is achieved by not relying on data obtained directly from the 3D sensor, but on data obtained by a mapping algorithm. In order to cluster the frontier points, we use the properties of the Octree environment representation, which allows easy analysis with different resolutions. The planner is tested in the simulation environment and in an outdoor test area with a UAV equipped with a lidar sensor. The results show the advantages of the approach.

\end{abstract}

\section{Introduction}

\begin{figure}[t!]
	\centering
	\includegraphics[width=1.0\columnwidth, trim={1cm 5cm 5cm 0.5cm}, clip]{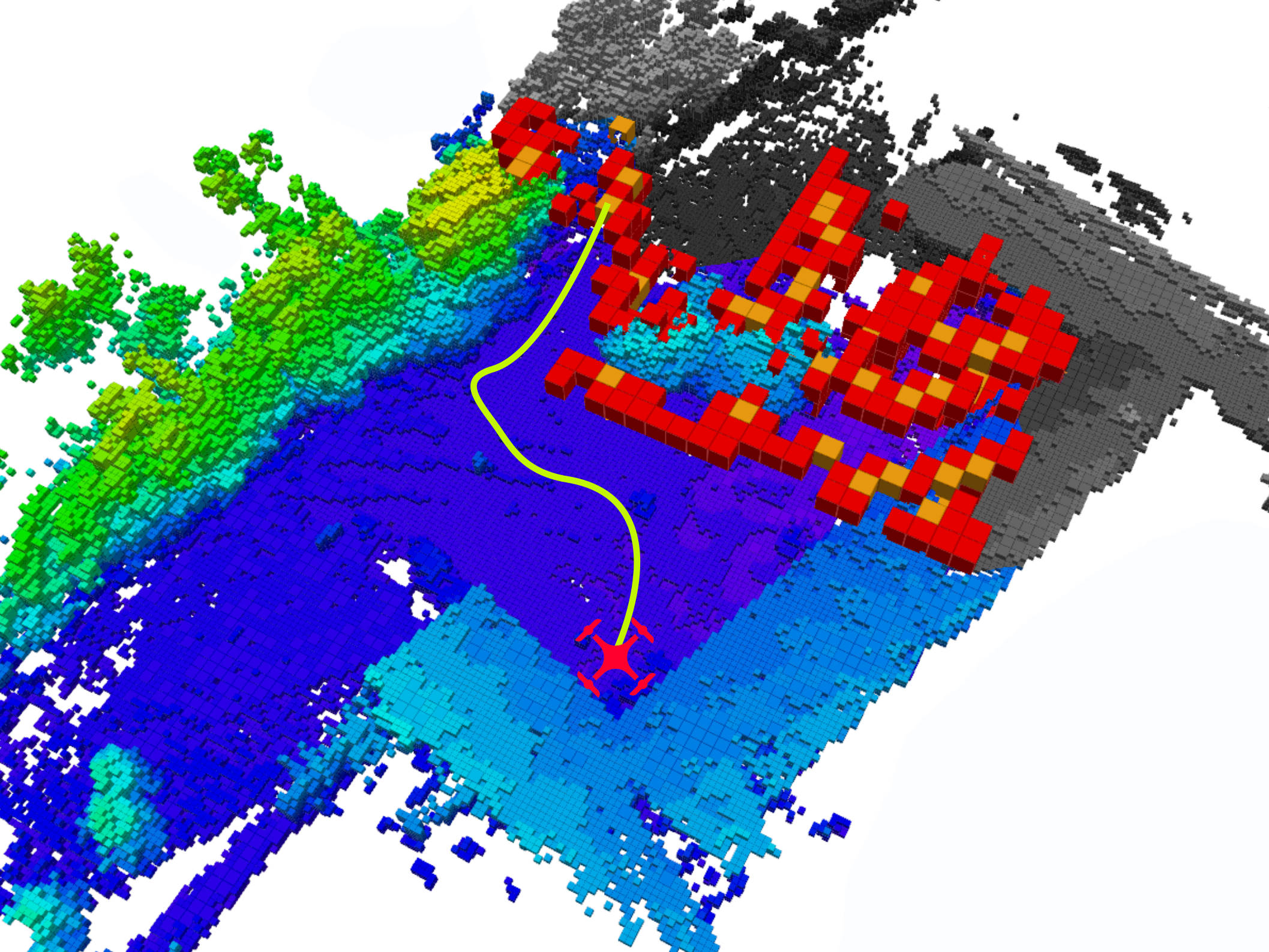}
	\caption{A snapshot of the proposed method in action. Grey and white parts of the OctoMap represent the unexplored and green/blue explored environment. Red voxels represent the frontier and yellow voxels are centroids of frontier point clusters. A UAV explores the environment by planning a trajectory towards the selected cluster centroid.}
	\label{fig:octomap_frontier}
\end{figure}

An autonomous exploration and mapping process is one of the fundamental tasks of robotics. Typical exploration methods are based on frontiers \cite{Yamauchi1997} and are used in both 2D and 3D space. In contrast to 2D exploration and mapping strategies, the mapping of large environments in 3D requires a high memory and computational effort. Therefore the fastest possible generation of a complete 3D map and autonomous navigation of a robot through the map is a challenging task.   
The main objective of this paper is to develop a 3D exploration planner capable of meeting the above challenges. The paper focuses on large unknown environments where a robot should navigate autonomously without any a priori knowledge of the environment. The planner, which consists of a sequence of algorithms, acts as a decision making tool that guides the robot to the next exploration point. A snapshot of the proposed method in action is shown in Fig. \ref{fig:octomap_frontier}.

We use Google's Cartographer SLAM algorithm \cite{Hess2016} as a basis for a novel exploration planner. For detection of the frontier, which is the first step of the exploration planning procedure, we use submap point cloud from Cartograpaher, which has one point for each occupied submap cell. A submap is created from a sequence of sensor scans by scan matching, fusion with IMU and odometry, thus providing a more stable input for the frontier detection algorithm compared to published exploration methods (\cite{Zhu2015}, \cite{Mannucci2017}, \cite{Baiming2018}, \cite{Bircher2016}) which use raw 3D sensor readings.  

We convert the aforementioned submap point cloud into an OctoMap, which has become the standard in recent years due to its efficient memory and querying properties \cite{Hornung2013}. For a more efficient frontier detection, we exploit the structure of the OctoMap which allows us to easily query occupied voxels with different spatial resolutions. Points of the frontier are detected at the lowest OctoMap level (corresponding to smallest Octree voxels), resulting in  large number of points which would decrease the overall performance of the proposed algorithm. To mitigate this problem, we efficiently filter frontier points by changing the resolution level of the points in the OctoMap. Final frontier points processing is performed using the mean-shift clustering \cite{Comaniciu2002} and results in a significantly reduced number of frontier points to be considered in further steps. The best frontier point to be visited is determined by estimating the benefit of the information gathered by visiting a candidate frontier point. The exploration loop is closed with an autonomous navigation to the selected frontier point, using the map generated through the exploration for trajectory planning and localization. 

This exploration procedure encompasses several novel elements that make up the contribution of this paper: a) a submap-based frontier detection; b) efficient multi-resolution frontier refinement; c) the best frontier point selection based on potential information gain. The validity and increased performance of the proposed approach is demonstrated through extensive analysis of simulation and experimental results. In addition, we focus our efforts on making our data sets (source, maps) available for comparison with other research approaches.

In Section \ref{sec:related} we give an overview of the state-of-the-art of 3D exploration methods and position our work in relation to them.  Section \ref{sec:proposed} is the core of the paper and contains the details of the planner. The results of the simulations and experiments performed with a UAV and their analysis, are presented in Sections \ref{sec:simulation} and \ref{sec:experiment}. The paper ends with a conclusion in Section \ref{sec:conclusion}.

\section{Related work}
\label{sec:related}

There is a wealth of earlier work related to autonomous exploration, especially for 2D, but more recently also for 3D environments. The approaches can be roughly divided into frontier-based and next best view-based approaches, 
even though there is a significant overlap between categories. In this section we give an overview of techniques from each category, with a focus on selected frontier-based approaches for 3D environment such as the one proposed in this paper. 

Characteristic to frontier-based approaches is exploration by approaching a selected point on the frontier between the explored and unexplored environment. This idea was first introduced by Yamauchi in \cite{Yamauchi1997} and tested in a 2D environment. The simplest approach to 3D exploration is to use 2D frontier- based exploration with 3D maps at different heights (sometimes called 2.5D approaches) (\cite{Alkhawaldah2012}, \cite{Bachrach2009}). A complete frontier-based solution for 3D environments is developed in \cite{Zhu2015} and \cite{Mannucci2017}, and these approaches are described in more detail later in this section. 

Next best view-based (NBV) approaches aim to determine a (minimal) sequence of robot (sensor) viewpoints in the environment to be visited until the entire search space is explored. Potential viewpoints are usually sampled, e.g. near the frontier or randomly. Then the viewpoints are checked for the potential information gain and the next best viewpoint is assigned. One of the first NBV methods is presented in \cite{GonzalezBanos2002} and then extended in \cite{Joho2007}, \cite{Bircher2016}, \cite{Baiming2018} and others. In \cite{Bircher2016} the authors use an RRT-based search to direct a UAV to the unexplored region. The method showed good scaling properties and the ability to handle large spaces, but due to the characteristics of the RRT algorithms, the total exploration time could be much higher for some environments. The exploration times are later improved in \cite{Baiming2018}. Often NBV approaches are used to build a 3D object without any a priori information, as in \cite{VasquezGomez2014} and \cite{Dornhege2013}.






Our approach was inspired by that of Zhu et al. \cite{Zhu2015}, an exploration tool called 3D-FBET. It is a frontier-based tool that is performed in three phases, similar to those presented in this paper. The phases are 3D mapping, frontiers detection in combination with a clustering algorithm, and the selection of the best frontier. Through experimental evaluation on different environments 3D-FBET showed several shortcomings. First, because the frontier detection is based on a subset of altered voxels (generated from camera point cloud), which is highly variable, the obtained frontiers were noisy and not reliable. Furthermore, the resulting frontier presented only a local view and the clustering was not adapted to the environment. These problems led to a higher total exploration time. The authors provide the source code and the duration analysis for each phase, which facilitates comparison with the new approaches. 

We extend this approach to recognize not only local but also global frontiers, similar to Mannucci et. al. \cite{Mannucci2017}. Mannucci proposed a 3D exploration with two OctoMaps and two frontiers (local and global) with different resolutions. Global frontiers are assigned when the set of local frontiers is empty. Manucci evaluates the best frontier using cost-utility approach, similar to \cite{Burgard2005}. Since maintaining two OctoMaps is a resource-intensive task, we use the properties of OctoMaps and implement a solution with multiple resolutions in a single OctoMap. 

Our 3D frontier detection is motivated by a dense 2D frontier method presented by Orsulic (\cite{Orsulic2019}), which has achieved good results in terms of wall time per frontier update. Together with multi-resolution clustering and appropriate target point selection, we are constructing a novel 3D exploration planner that accelerates the 3D exploration process. Our planner is able to run online and on board a robot with limited resources. The results are shown in simulations and experiments and data sets are provided for further use.


	

\section{Proposed approach}
\label{sec:proposed}


The problem of autonomous exploration of either indoor or outdoor unknown 3D space $V \subset \mathbb{R}^{3}$ has the ultimate goal to create a 3D map of the environment. 

As a basis for our algortihm, we use an OctoMap, a hierarchical volumetric 3D representation of the environment. Each cube of the OctoMap is called a voxel (cell) $v$, which can be \textit{free}, \textit{occupied} or \textit{unknown}. Free voxels form free space $V_{free} \subset V$, occupied voxels occupied space $V_{occ} \subset V$ and unknown voxels unknown space $V_{un} \subset V$. The entire space is a union of the three subspaces $V \equiv V_{free}. \cup V_{occ} \cup V_{un}$. 

The problem addressed in this paper is how to design an exploration planner for a robot, i.e. how to select suitable waypoints for the robot at appropriate times, with the aim of traversing unknown parts of the environment in the shortest possible time and with the least possible energy expenditure. The exploration is considered complete when $V_{un} = \emptyset$. The approach we are pursuing is a frontier-based exploration strategy, where the goal is to increase the overall knowledge of the environment by directing the robot to the \textit{frontier point} with the best trade-off between benefit and cost. 

\begin{figure}[t!]
	\centering
	\includegraphics[width=1.0\columnwidth]{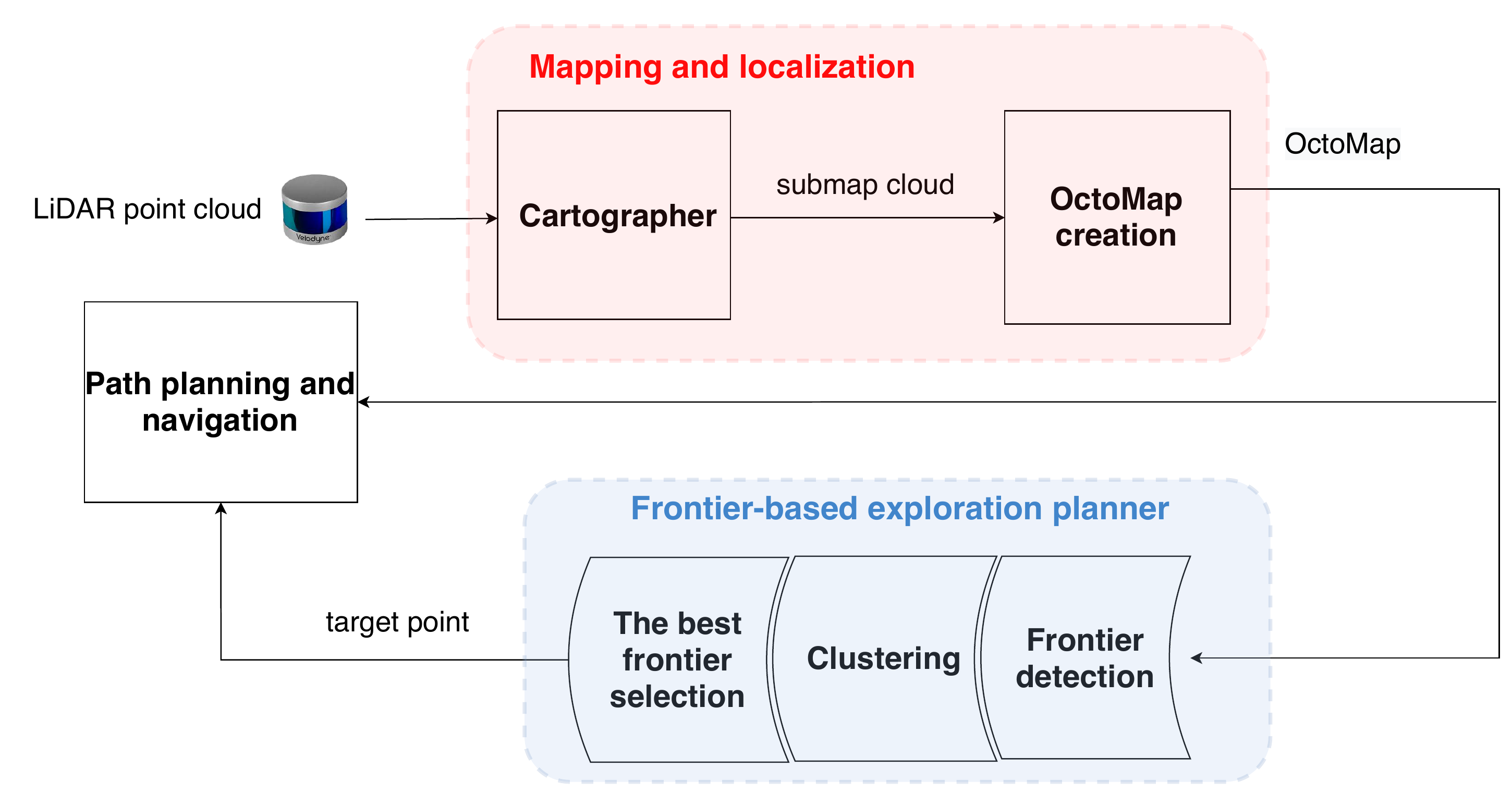}
	\caption{Overall schematic diagram of autonomous 3D frontier- based exploration. The Cartographer SLAM creates 3D submaps, which are an input for the OctoMap generation module. Frontier detection, clustering and the module for selecting the best frontier voxels form the proposed 3D exploration planner (highlighted in blue). The best frontier voxel represents a target towards which the robot navigates, taking into account the OctoMap created so far.}
	\label{fig:diagram}
\end{figure}

In this work, the exploration is performed with an unmanned aerial vehicle (UAV) that has no prior knowledge of the environment. Although the concepts are explained with an UAV in mind, the same approach can be used with other types of autonomous robots. For map generation and localization we assume a suitable algorithm for simultaneous localization and mapping (SLAM) exists, which requires an appropriate sensing system, e.g. a laser scanner or camera. An OctoMap is generated using the SLAM algorithm and is used for both frontier detection and collision-free navigation of the UAV. We also assume that a suitable path planning and following algorithms are available for the UAV. Overview of the proposed system is given in Fig. \ref{fig:diagram}

\subsection{Frontier detection}
\label{subsection:frontier_detection}

A frontier, $F$, can be defined as a set of voxels $v_{f}$ with the following property:

\begin{equation}
    F = \{ v_f \in V_{free} : \exists neighbor(v_f) \in V_{un}\}. 
    \label{eq:front_def}
\end{equation}
In other words, frontier consists of free voxels with at least one unknown neighbor. Center of a frontier voxel is often referred to as frontier point. Since space $V$ is bounded, once the exploration is over frontier becomes empty, $F = \emptyset$, which leads to $V_{free} \cup V_{occ} = V$. 

As already mentioned, OctoMap used for frontier detection is generated using Cartographer submaps. Submap $m^{i}_{s}$ ($i^{th}$ submap) is built using the last $N_s$ consecutive LiDAR scans $S$, and matched against past IMU and odometry data: 
\begin{equation}
m^i_{s} = f(S_{(i-1)N_s}, \ldots, S_{iN_s}, IMU, Odometry),
\end{equation}
where $S_k$ denotes $k^{th}$ LiDAR scan. Function $f$ stands for a nonlinear optimization that aligns each successive scan against a submap being built. When the predetermined fixed number of scans $N_s$ is inserted into a submap, it is marked as completed. Note that the size of a submap is adjustable, which makes the entire exploration process more robust. The map $m$ can be created by joining all past submaps together:
\begin{equation}
m^i = f(m^1_s, \ldots, m^i_s).
\end{equation}
Both map $m$ and submaps $m_s$ are in form of a 3D occupancy grid. A format much more suitable for path planning and other operations are octrees, so instead of building a 3D occupancy grid map $m$ we build an OctoMap $O$ using the OctoMap generation software \cite{Hornung2013}. Namely, from each completed submap $m^i_s$ we calculate a submap cloud $m^i_{sc}$ by adding a point in the centre of each occupied voxel of $m^i_s$. OctoMap $O^i$ is then created from all the past submap clouds:
\begin{equation}
O^i = f(m^1_{sc}, \ldots, m^i_{sc}).
\end{equation}

\begin{figure}[t!]
    \centering\includegraphics[width=1.0\columnwidth]{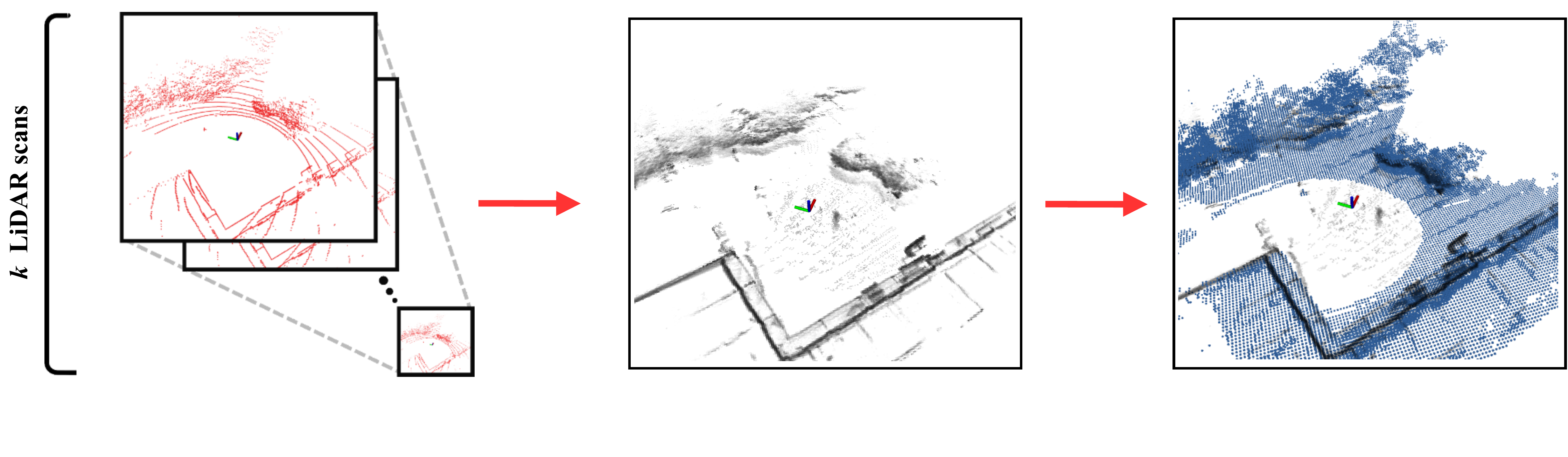}
	\caption{Creating a submap cloud from $N_s$ LiDAR scans (red) that are  matched against a submap $m_s$ (black) and a submap cloud $m_{sc}$ is created at the end (blue).}
  \label{fig:k_scans}
\end{figure}

The process of creating submap clouds from sensor scans is shown in Fig. \ref{fig:k_scans}. Due to optimizing the $N_s$ laser scans to form a submap, submap clouds provide a more stable input for Octomap generation compared to raw scans from the sensor, which enables more reliable detection of frontier points.

Similar to FBET \cite{Zhu2015}, our algorithm extracts frontiers incrementally. Each time a new submap is created the OctoMap is updated and a new frontier detection cycle is started. Our approach combines local and global frontiers, similar to Manucci \cite{Mannucci2017}. Local frontier $F_l$ is derived directly from the OctoMap and updated with each new-coming submap cloud, as follows:
$$F^i_l = O^i - O^{i-1},$$
where subtraction of two OctoMaps keeps only the changed voxels with respect to the older OctoMap. Global frontier $F_g$ is a union of all past local frontiers, updated in each iteration and filtered to exclude voxels which have been discovered in the mean-time: 
\begin{equation}
    F^i_g = \cup_{k=1}^{i} F^k_{l}.
\end{equation}

There is usually a large number of voxels in the global frontier (referred to only as frontier from now on) and their evaluation is expensive in view of the computing effort involved. Therefore, we cluster frontier voxels using both multi-resolution frontier and mean-shift clustering algorithms. This procedure leads to multiple clusters whose geometric central voxels are potential exploration targets.

\subsection{Multi-resolution frontier clustering}
\begin{figure}[t!]
		\centering
		\includegraphics[width=1.0\columnwidth]{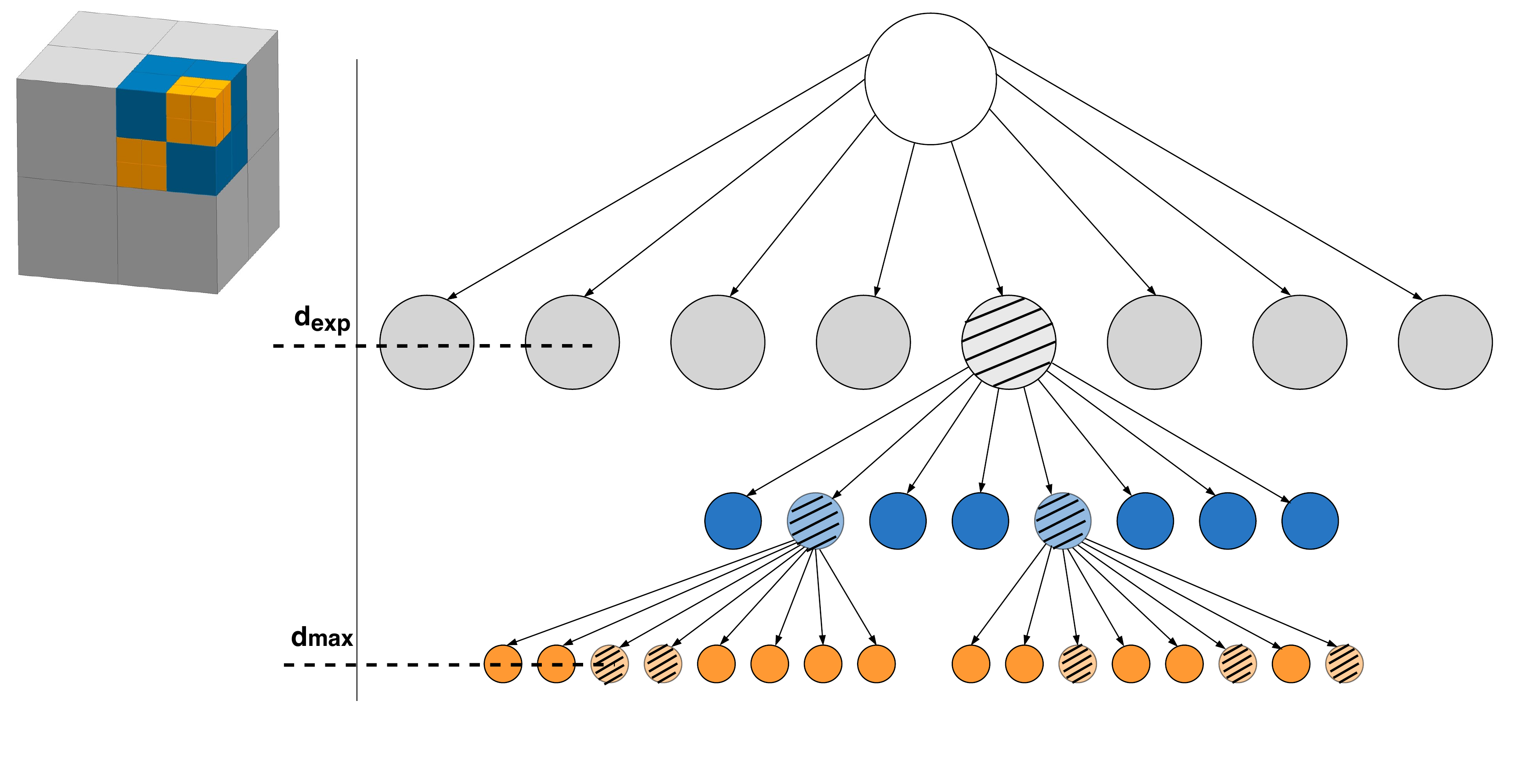}
		\caption{The structure of an octree and the cube shaped space it represents. An instance of frontier voxels at $d_{max}$ and their parents on the desired exploration depth $d_{exp}$.}
	\label{fig:multi-resolution}
\end{figure}

In this section we exploit the Octree structure of the OctoMap to perform initial clustering of frontier voxels. The Octree is a tree structure consisting of a node, or \textit{OcTreeNode}, which has eight children (Fig. \ref{fig:multi-resolution}). The children are also OcTreeNodes, which means that an Octree recursively divides the volume.  The  maximum depth of  the  OctoMap is $d_{max}=16$ \cite{Hornung2013}. The size of the voxel at this maximum level determines the level of detail of the OctoMap and is denoted with $r_{max}$.

In this work, the OctoMap is used for path planning, navigation and exploration. In the path planning and navigation process it is crucial to keep the map resolution as high as possible, that is, we use the OctoMap level $d_{max}$ for UAV navigation. In this work the frontier is being detected at OctoMap level $d_{max}$, but in general we can choose other levels for initial frontier detection. When making this decision, one should consider the expected structure of the environment, as calculation on lower levels may artificially close corridors or narrow paths through the environment. The trade-off for calculation of frontier on $d_{max}$ is large number of frontier points, which can cause unnecessary consumption of computational resources in later stages of the exploration planning procedure, especially if we focus on large outdoor environments. For that reason we aim to decrease the number of frontier points for future consideration while exploiting OctoMap  multi-resolution properties for efficient frontier clustering. 

We define the desired exploration level $d_{exp}$ and the corresponding exploration voxel size $r_{exp}$ based on the characteristcs of the environment. If we expect more open areas, $d_{exp}$ can be lower and $r_{exp}$ can be larger.
Frontier points clustered to level $d_{exp}$ are denoted as $F_{exp}$ and determined as follows. Let us consider only four depth levels (as shown in Fig. \ref{fig:multi-resolution}), and let frontier detection level be $d_{max} = 4$ and the desired exploration level be $d_{exp} =2$. Then our goal is to find frontier parent OcTreeNodes from depth $d_{exp}$ that are parents to the known frontier voxels $v_f$ from $d_{max}$. A general expression for determination of frontier points clustered to the exploration level in iteration $j$ (parent frontier voxels) $F_{exp}$ is:
\begin{equation}
    F^j_{exp} = \{v^j_{exp}: v^j_{exp} = parent (v^j_{f}) \text{ at } d_{exp}\},  \forall v^j_{f} \in F^j_g.
\end{equation}
We use superscript $j$ in the previous equation to emphasise that the process of clustering is not performed for each newly built OctoMap, but only when the UAV reaches the $j-th$ commanded waypoint, which was generated by previous $(j-1)$ exploration planner iteration. The frontier is updated after each $N_s$ lidar scans, i.e. when a new submap is created, because we might miss an important frontier update if we do it less frequently. The described multi-resolution clustering algorithm is fast and robust, easy to implement and suitable for different map resolutions and exploration depths. It can be applied to small and large areas. In our approach we combine it with the mean-shift clustering algorithm because we want to direct a robot into the area where frontiers are denser.  


\subsection{Mean-shift frontier clustering}
\label{subsection:mean-shift}

Clustering algorithms are often used in an exploration process (union-find algorithm, depth-first algorithm or flood fill clustering algorithm are used in \cite{Dornhege2011}, \cite{Zhu2015} and \cite{Mannucci2017} respectively). In contrast to state-of-the-art approaches, we use mean-shift clustering algorithm applied to 3D points. The mean-shift was first proposed by Fukunaga and Hostetler \cite{Fukunaga1975} and requires no assumptions on the form of the distribution or the number of clusters (compared to for example \textit{K-means} \cite{Duda2001}).

The input into the mean-shift clustering are parent frontier voxels obtained in the previous step, $F_{exp}$. The output of the mean-shift clustering are frontier voxels which are candidates for becoming a next waypoint for the UAV and are denoted as $F_c$. 

The computationally most complex component of the mean-shift procedure corresponds to the identification of the neighbours of a point in 3D space (as defined by the kernel and its bandwidth). To make the mean shift algorithm work in real time, along with the reduction in the number of global frontiers by multi-resolution frontier clustering, we carefully selected an appropriate bandwidth to balance between computation time and the desired outcome with respect to the size of the environment and resolution $r_{exp}$.

\subsection{Best frontier voxel selection}

To evaluate which of the voxels in $F_c$ could result in a faster exploration of the environment, we define \textit{total gain} of every candidate $v_c \in F_c$ using the following function similar to one proposed in \cite{GonzalezBanos2002}:

\begin{equation}
	G(v_c) = \frac{I(v_c)}{e^{\lambda L(p_i, p_{vc})}},
	\label{eq:totalgain}
\end{equation} 
where $\lambda$ is a positive constant, $L(p_i ,p_{vc})$ is the distance between robot's current position $p_i$ and the position of the candidate $v_c$, while $I(v_c)$ is an \textit{information gain} i.e. a measure of the unexplored region of the environment that is potentially visible from $v_c$. 
The estimated distance is approximated using Euclidean distance between the robot position $\boldsymbol{p_{i}}$ and the position of the candidate (voxel center), $\boldsymbol{p_{v_c}}$:

\begin{equation}
 L(p_i, p_{v_c})= \lVert p_{i}-p_{vc}\rVert. 
\end{equation}

The information gain $I(v_c)$ is defined as the share of unknown voxels in a cube placed around $v_c$. Size of the cube is defined with respect to the range of the used sensor. Often the information gain is estimated using a ray tracing algorithm and a real sensor field of view instead of using a cube-based approximation. By using the proposed simplification, the high calculation effort required by ray tracing is avoided.

The constant $\lambda$ weights the importance of robot motion cost against the expected information gain. A small $\lambda$ gives priority to the information gain, while $\lambda \rightarrow \infty $ means that the motion is so expensive that only $v_c$ near the robot is selected.

We can calculate the value of $\lambda$ to be used based on relative importance we set on motion cost and information gain. If we have two candidates $v_{c1}$, $v_{c2}$ and their information gains $I(v_{c1})$, $I(v_{c2})$, we can choose $\lambda$ as follows: 

\begin{equation}
	\lambda = \frac{\ln (\frac{I(v_{c2})}{I(v_{c1})})}{L(p_i, v_{c2}) - L(p_i, v_{c1})}.	
\end{equation} 

In this way, it is easy to set the ratio between the desired information gain and the distance with respect to the desired behavior of the system. For instance, if we want to prefer twice less information gain only if $L(p_i, v_{c2}) - L(p_i, v_{c1}) > 5m$, we set $\lambda$ to 0.1386. 

Finally, the best frontier voxel is one that maximizes total information gain $G(v_c)$:

\begin{equation}
v_{bf} = \argmax_{v_c \in F_{C}} G(v_c).
\end{equation} 


As soon as the best frontier point is selected, it is forwarded to a path planner as a waypoint. A robot starts to follow the planned path and navigates to the best frontier point $v_{bf}$. For UAV control, we use a RRT-based path planner and trajectory following solution \cite{Arbanas2018}. New cycle of the procedure for determination of the best frontier is started after the previous waypoint is reached by the UAV, that is, the clusters and candidate frontier voxels $F_C$ are re-calculated. The exploration process is performed until the entire environment is explored and a complete map of the environment is created. 
\section{Simulation-based evaluation}
\label{sec:simulation}
The simulations are performed in the Gazebo environment using Robot Operating System (ROS) and a model of the \textit{Kopterworx} quadcopter, which is identical to the one used for experiments in the real world. The quadcopter is equipped with a Velodyne VLP-16 LiDAR sensor, whose maximum range is reduced to 20 m in simulations. We set the maximum velocity of the UAV to 0.8 m/s\footnote{Maximum velocity in simulation is set to velocity used in outdoor experiments, which is low for safety reasons} and run two scenarios with different sizes of the environment to be explored, and analyze the results. All tests are processed on the computer with an Intel i7 8550U processor.

\subsection{House Exploration Scenario}

\begin{figure}[t]
	\centering
	\begin{minipage}{0.38\columnwidth}
		\centering
		\includegraphics[width=1\columnwidth]{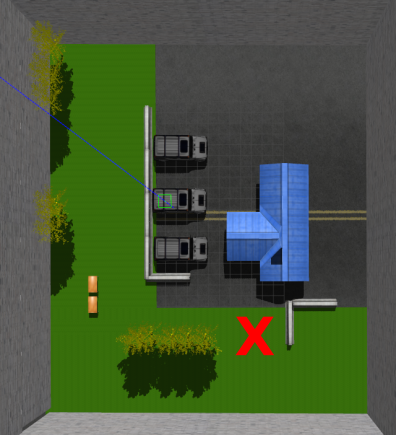}
		\subcaption{}
		\label{fig:a}
	\end{minipage}
	\begin{minipage}{0.56\columnwidth}
		\centering
		\includegraphics[width=1.2\columnwidth]{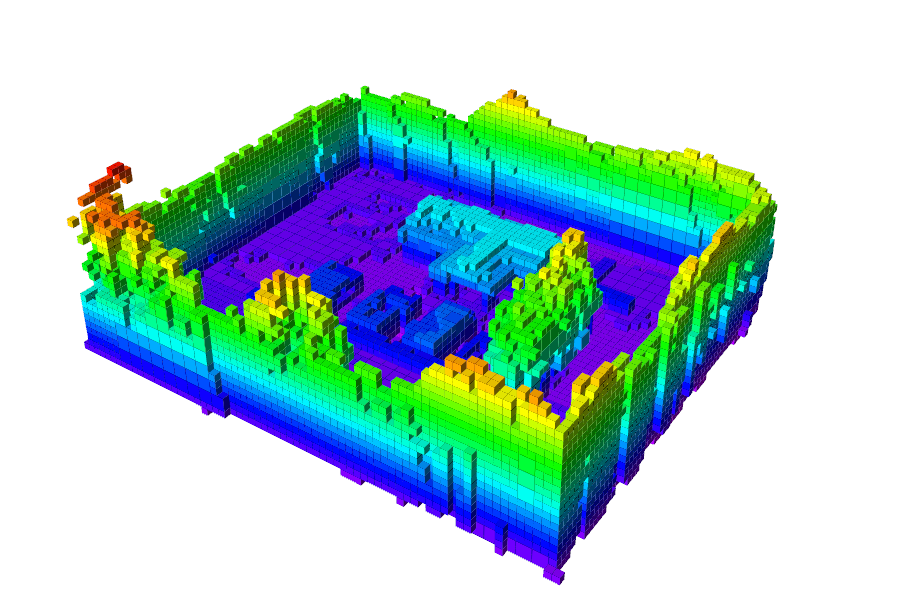}
		\subcaption{}
		\label{fig:b}
	\end{minipage}
	\caption{House exploration scenario. (a) Gazebo world. (b) OctoMap generated during exploration.}
	\label{fig: house_scenario}
\end{figure}

\begin{figure}[t!]
	\centering
	\includegraphics[width=1.0\columnwidth]{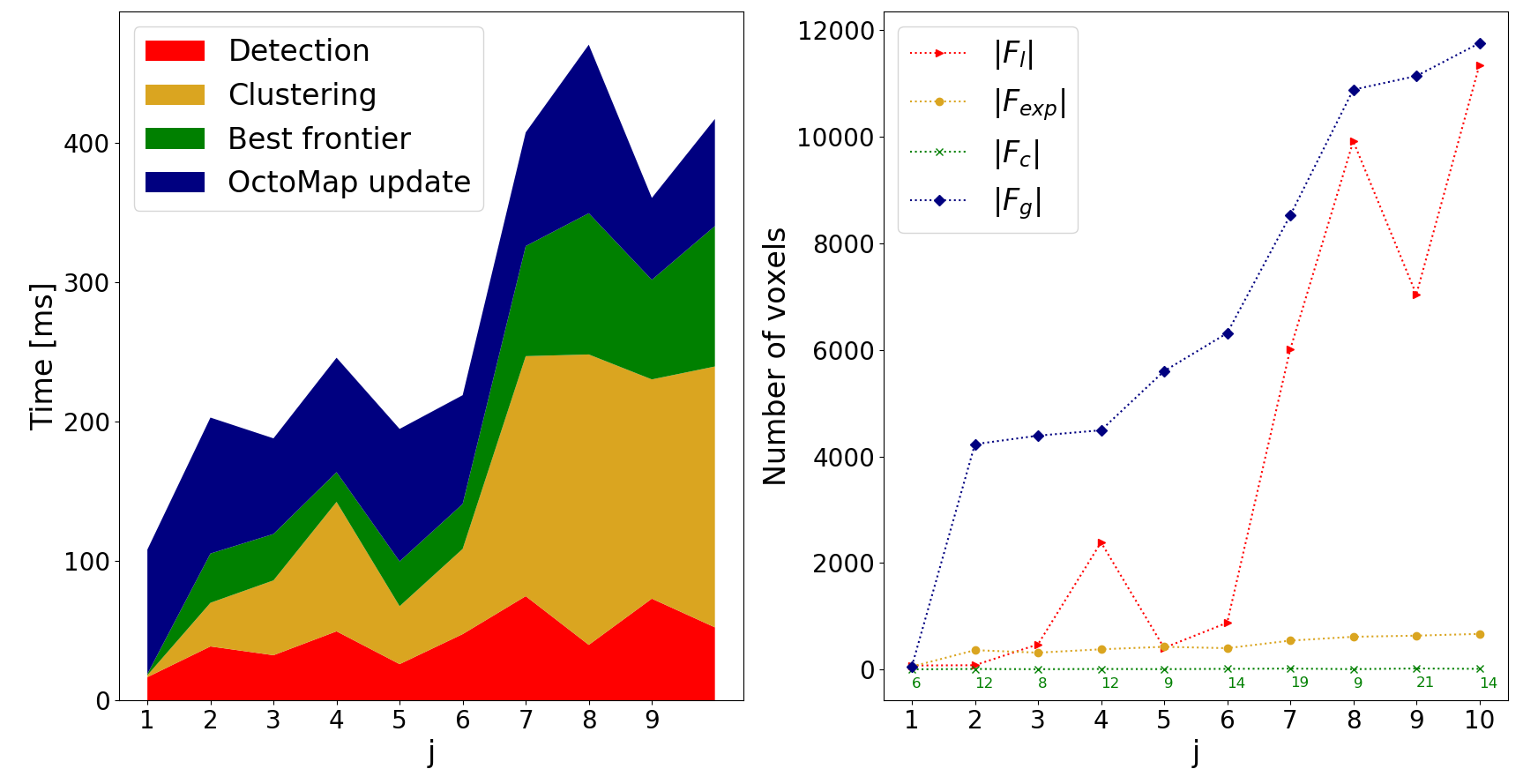}
	\caption{House scenario metrics for $r_{max} = 0.25$m, $r_{exp} = 1$m, bandwidth = 2.}
	\label{fig: house2_final2}
\end{figure}

The first scenario refers to a 30x40x5m space shown in Fig. \ref{fig: house_scenario}.
The vehicle starts from the marked position in the Gazebo world and navigates through the environment to explore the entire space. 
For the first scenario, the voxel size at the lowest resolution level is $r_{max}=0.25$m. For exploration we use level $d_{exp} = 14$, that is, voxel size $r_{exp} = 1$m and mean-shift bandwidth of 2. The results are shown in Fig. \ref{fig: house2_final2}.
The figure shows, for each iteration $j$ of waypoint calculation, the number of frontier voxels $|F_g|$, local frontier voxels $|F_l|$, number of parent frontier voxels at the exploration depth $|F_{exp}|$ and final target candidates, $|F_c|$. Furthermore, computation times for significant modules (OctoMap creation, frontier detection, clustering and best frontier selection) are also given. 
Note that clustering includes both multi-resolution and mean-shift algorithms, while detection takes into account time required to detect local frontiers and to update global frontiers.
These modules take part in total computation time, also calculated in each iteration. For this scenario, the average computation time is 0.197s with the standard deviation 0.062s, which shows that our planner is suitable for real-time performance. 

\begin{figure}[t!]
	\centering
	\includegraphics[width=0.75\columnwidth]{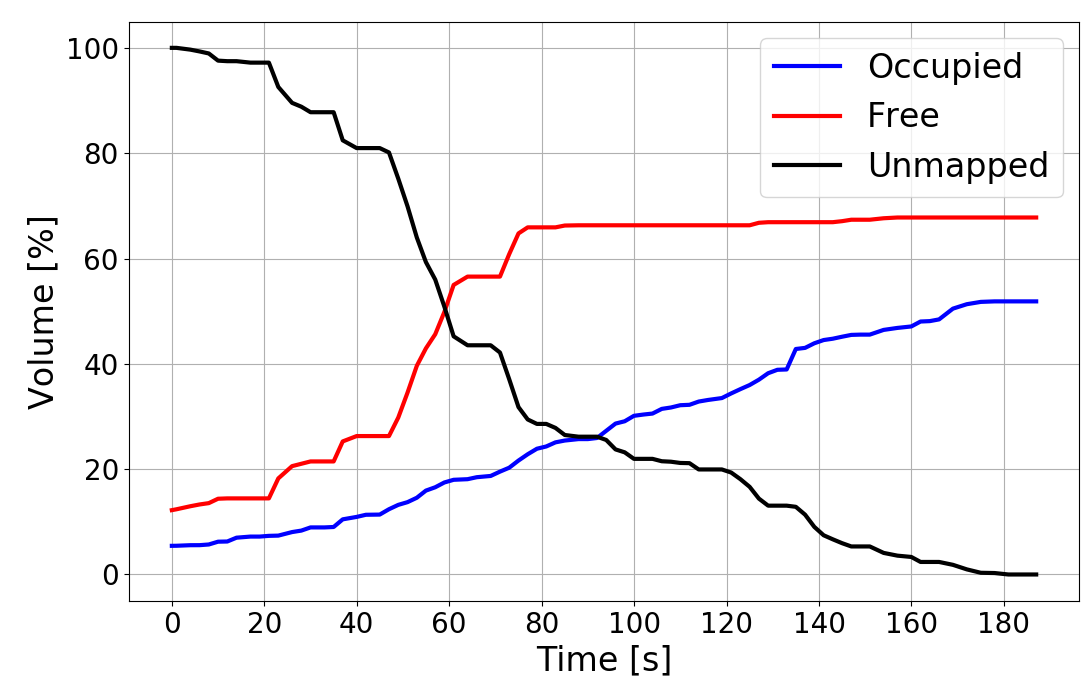}
	\caption{House scenario - the percentage of free, occupied and unmapped volumes in time.}
	\label{fig: volume_house}
\end{figure}

The percentage of free, occupied and unmapped volumes during the exploration is shown in Fig. \ref{fig: volume_house}. Note that the total exploration time here includes the computation and execution time as well as the RRT-based path planning time. The graph shows that we need less than 200s to explore the environment in the house scenario (Fig. \ref{fig: house_scenario} (b)).

\subsection{Large Exploration Scenario}
\begin{figure}[t!]
    \centering
	\begin{minipage}{0.45\columnwidth}
	   \centering
	   \includegraphics[width=\columnwidth]{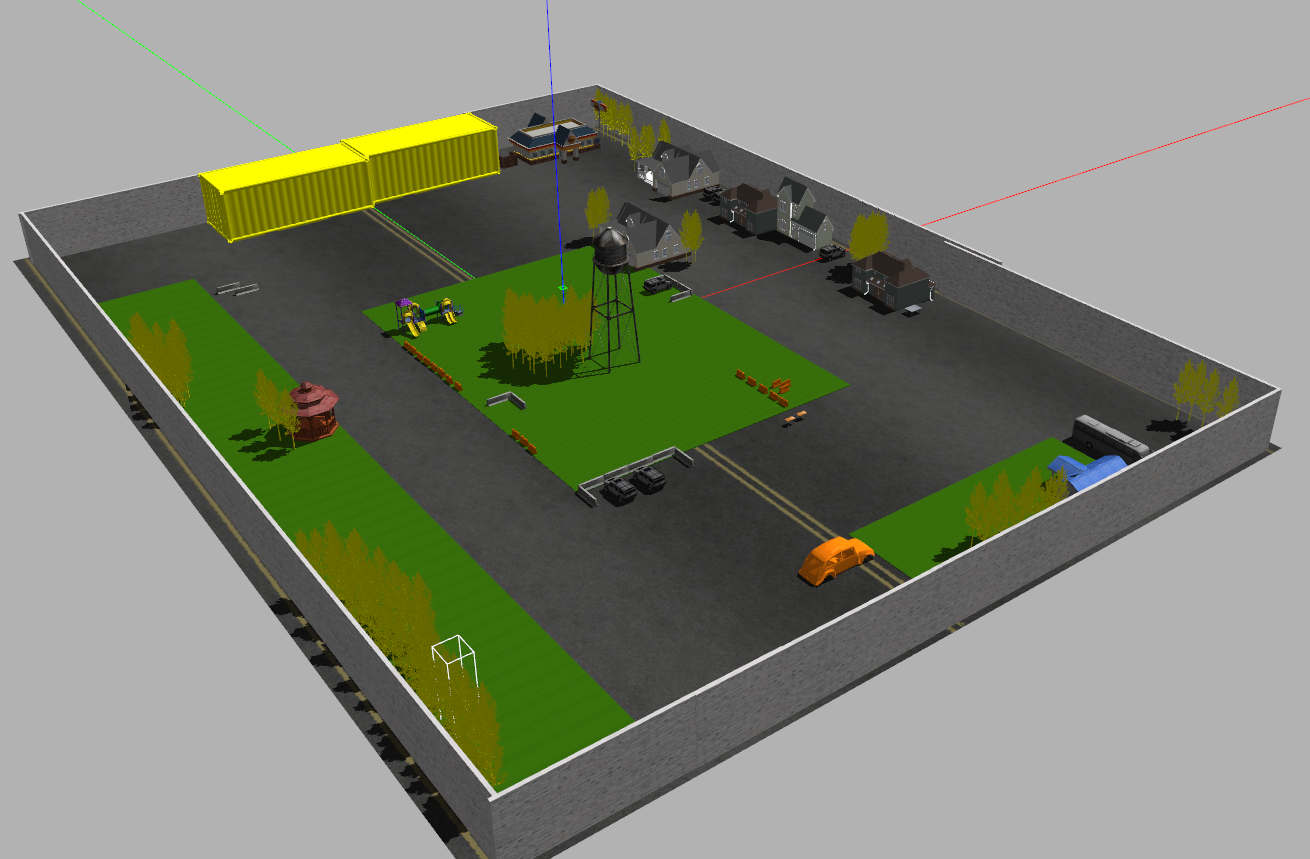}
	   \subcaption{}
	   \label{fig:a}
	\end{minipage}
\hfill 	
	\begin{minipage}{1\columnwidth}
	   \centering
	   \includegraphics[width=\columnwidth]{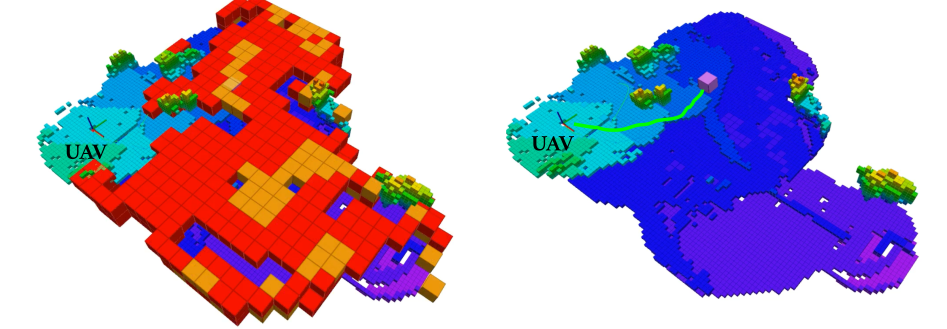}
	   \subcaption{}
	   \label{fig:b}
	\end{minipage}
\caption{Large exploration scenario. (a) Gazebo world (b) An instance of the exploration process. Global frontiers are marked red, candidates yellow while the best frontier voxel $v_{bf}$ is marked pink. The UAV planned a path (green) to the best frontier voxel (target).}
\label{fig: large_scenario}
\end{figure}

The second scenario refers to a 130x160x5m space, similar to the real world environment, shown in Fig. \ref{fig: large_scenario} (a).

\begin{figure}[t!]
	\centering
	\includegraphics[width=1.0\columnwidth]{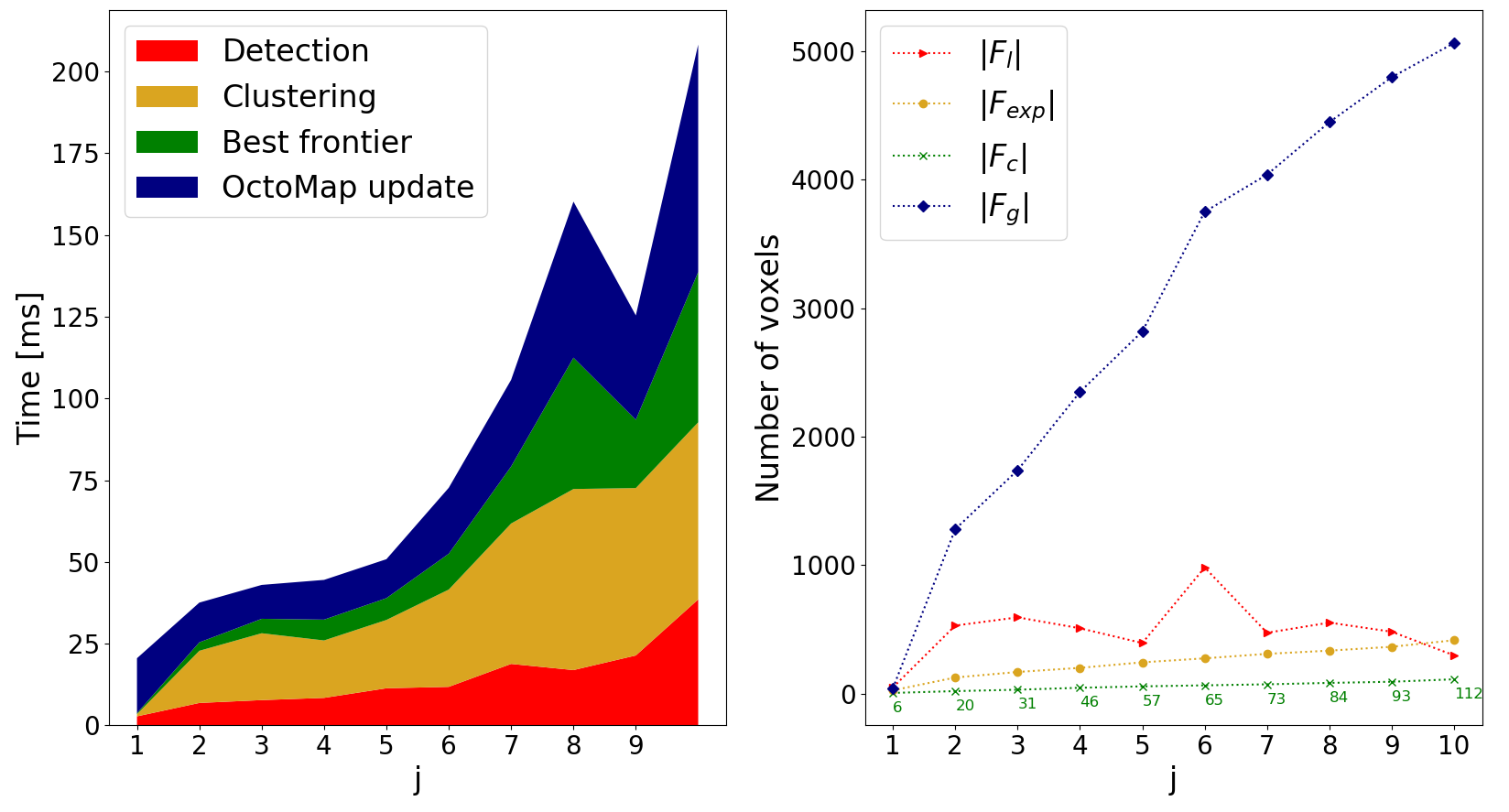}
	\caption{Large scenario metrics for $r_{max} = 0.5$m, $r_{exp} = 2$m, bandwidth = 2}
	\label{fig: large_final}
\end{figure}

In this scenario we set $r_{max}$ to 0.5 m, $d_{exp}$ to 14 and the bandwidth to 2. We can allow a lower resolution and exploration depth, as this scenario is larger than the house environment described above. 
An instance of the large scenario exploration with the mentioned parameters is shown in Fig. \ref{fig: large_scenario} (b). Global frontiers (red) are clustered, resulting in candidates (yellow). The path is computed and UAV navigates to the best frontier voxel (pink).  
The number of frontiers and the corresponding computation time during a single run are given in Fig. \ref{fig: large_final}. 

\begin{figure}[t!]
	\centering
	\includegraphics[width=0.8\columnwidth]{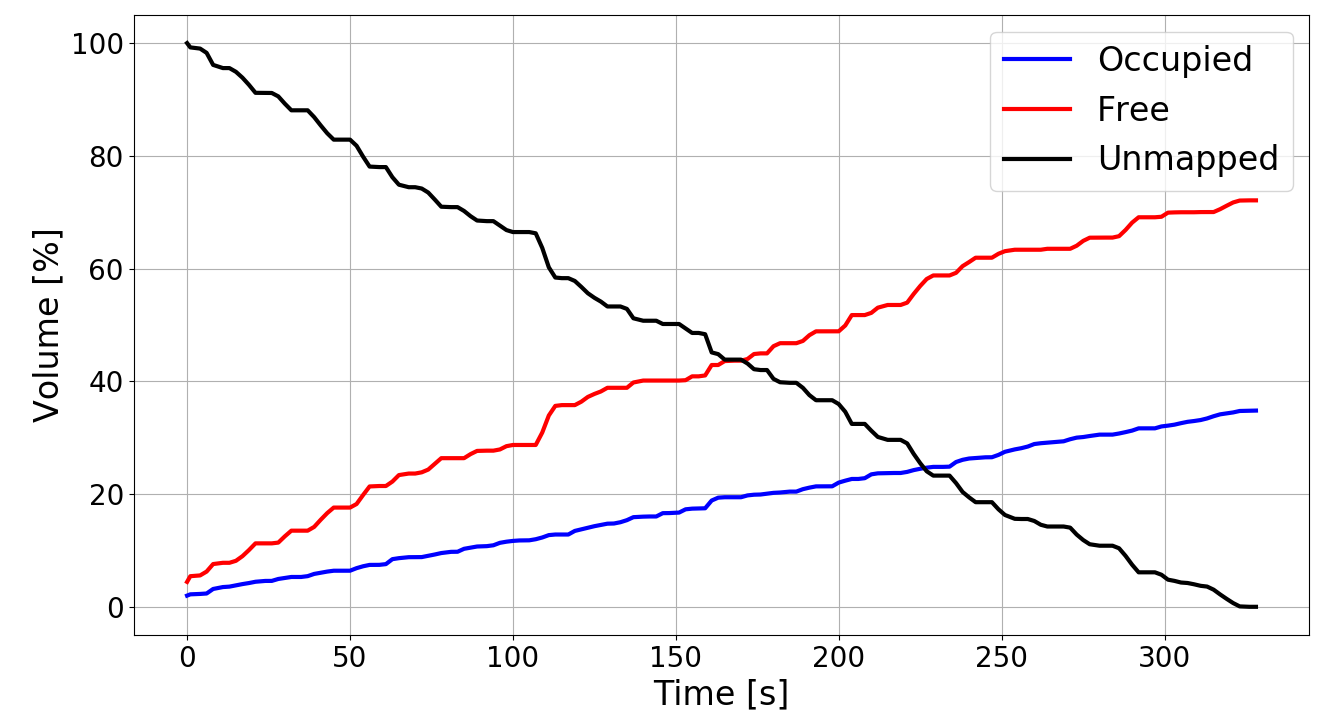}
	\caption{Large scenario - The  percentage  of  free,  occupied  and  unmapped volumes in time}
	\label{fig: volume_large}
\end{figure}

Even if the environment is larger, our planner needs on average only 0.095s for the entire calculation time with a standard deviation of 0.058s. In other words, our planner is also suitable for larger environments where the resolution may be lower. 
As shown in Fig. \ref{fig: volume_large}, the total exploration time is about 350s. Note that the slope of the curve is different from that shown in Fig.  \ref{fig: volume_house}. There are no sudden jumps in the explored volume because the environment is larger, but the LiDAR range is the same.

\subsection{Simulation results discussion}
Based on the simulation results shown in the last subsections, some general conclusions can be drawn which can be used in the design of an exploration system. First, frontier detection computation time and clustering time increase directly with the size of the global frontier, and we can vary the size of the global frontier by setting different values of $r_{max}$. Next, the computation time for the best frontier point depends directly on the number of candidate voxels $F_c$, and we can vary the number of candidate voxels for any $r_{max}$ by changing the exploration level $d_{exp}$ and the bandwidth parameter of the mean shift. As expected, the OctoMap update time does not depend on the size of the frontier. 
The averages of the total computation time required for one exploration planner iteration (0.197s, 0.095s) allow the process to run even more frequently than in the current solution.  The total computation time has approximately linear relation to the size of frontier, and the absolute values are suitable for applications under consideration, so we can say that the approach scales well with the increase in resolution and number of frontier points. Simulations were performed for different initial UAV positions and similar numerical values were obtained.
Direct comparison of showed simulation results with other state-of-the-art approaches is difficult, due to different setups. We mention the numerical results from \cite{Mannucci2017} where authors show results for an arena 100x80x7m, $v_{max}=1$m/s, $r_{max}= 1.5$m, stereo camera with a limited field of view and achieve a total exploration time of 1424s using a single robot. Computation times are not given. Even though these numbers are better in our approach, we need to test solutions in the same setting to make further conclusions.

\section{Experimental evaluation}
\label{sec:experiment}

\subsection{Setup}
\begin{figure}[t!]
	\centering
	\includegraphics[width=0.85\columnwidth]{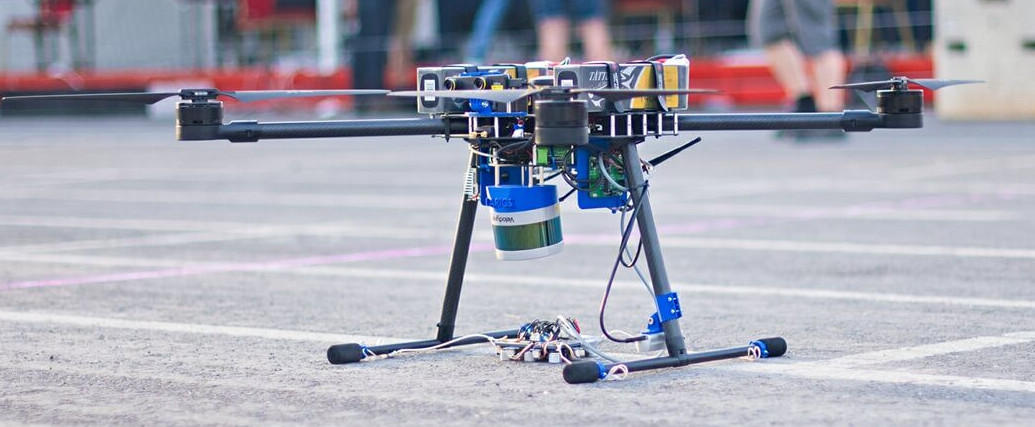}
	\caption{A custom built quadcopter equipped with a Velodyne VLP-16 LiDAR sensor}
	\label{fig:uav}
\end{figure}

For our outdoor experimental analysis, we use a custom built quadcopter (Fig. \ref{fig:uav}) assembled by \textit{Kopterworx}. The UAV features four T-motors P60 KV170 motors attached to a carbon fiber frame. The dimensions of the UAV are $1.2m \times 1.2m \times 0.45m$, which makes it a relatively large UAV suitable for outdoor environments. The total flight time of the UAV is around $30min$ with mass of $m=9.5kg$, including batteries, electronics and sensory apparatus. The \textit{Pixhawk 2.1} flight controller unit is attached to the center of the UAV body, and it is responsible for low-level attitude control of the vehicle. Furthermore, we equipped the UAV with \textit{Intel NUC} on-board computer for collecting and processing sensory data. The on-board computer runs \textit{Linux Ubuntu 18.04} with \textit{ROS Melodic} framework that communicates with the autopilot through a serial interface. The UAV is equipped with a Velodyne VLP-16 LiDAR sensor with a maximum range of 100 m. The maximum velocity of the UAV is limited to 0.8 m/s with a maximum acceleration of 0.5 m/s$^2$.
\subsection{Results and discussion}
\begin{figure}[t!]
	\centering
	\includegraphics[width=0.9\columnwidth]{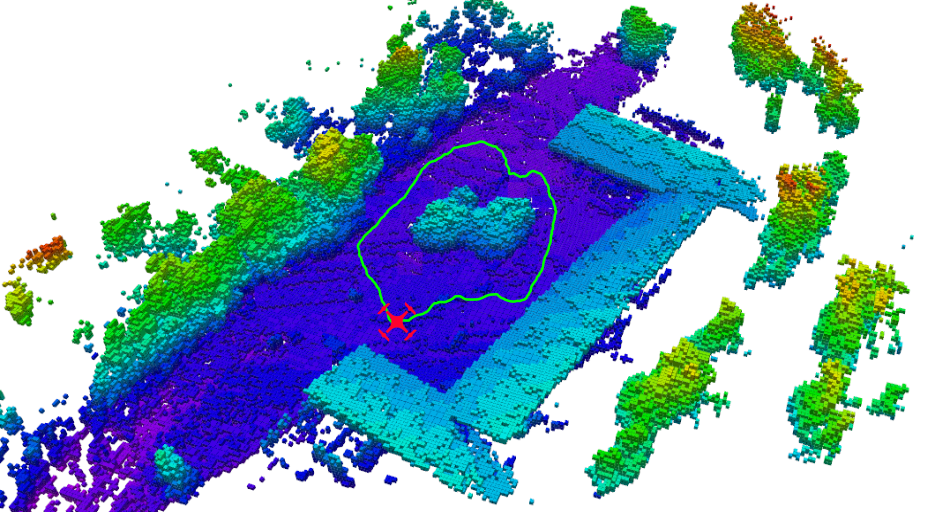}
	\caption{The OctoMap created during the exploration of the real world scenario with the path traversed by the UAV during exploration.}
	\label{fig: path_borongaj}
\end{figure}

\begin{figure}[t!]
	\centering
	\includegraphics[width=1\columnwidth]{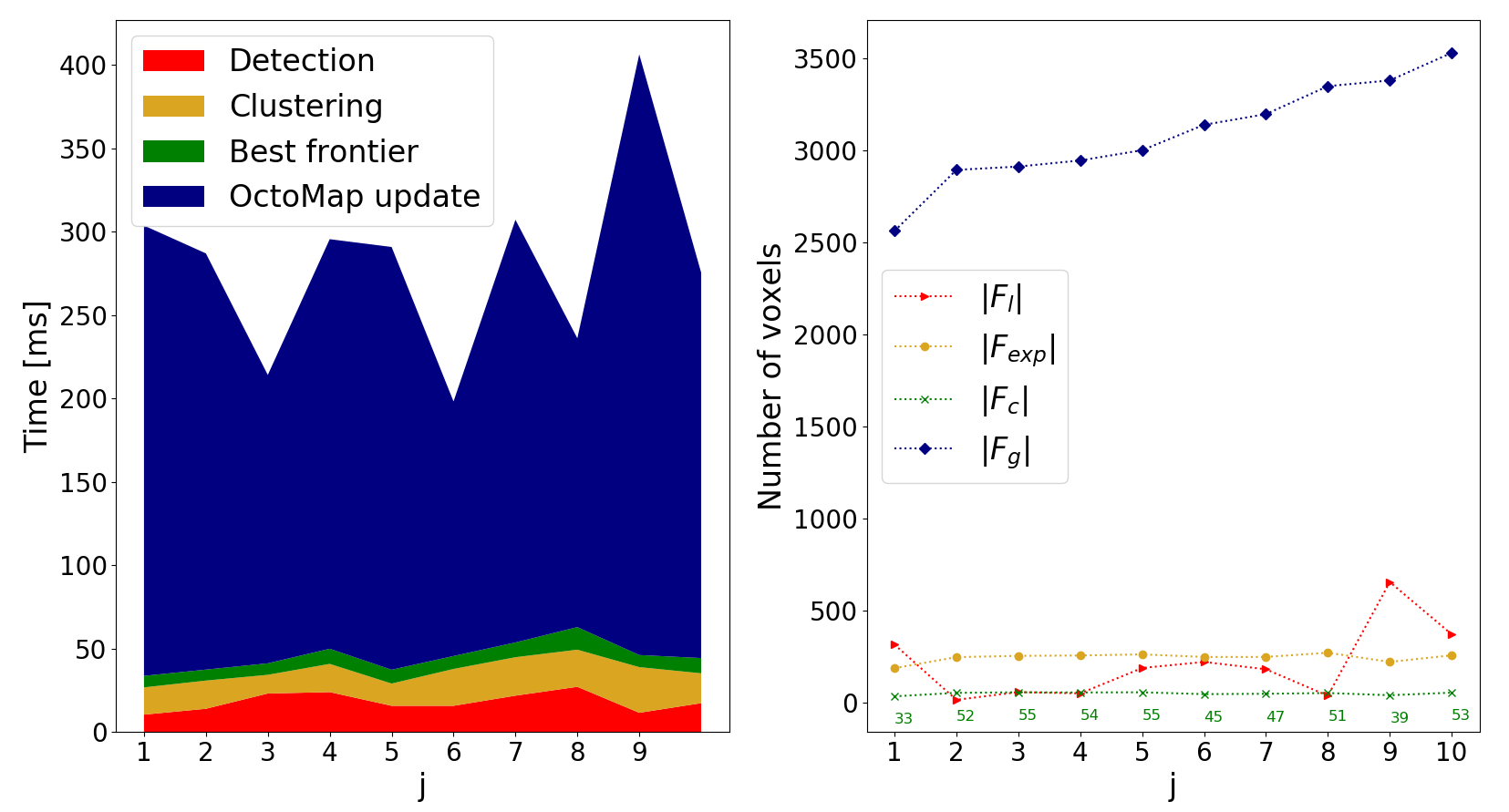}
	\caption{Outdoor scenario metrics for $r_{max}$ = 0.5 m, $r_{exp}=2$, bandwidth = 2}
	\label{fig:borongaj}
\end{figure}

\begin{figure}[t!]
	\centering
	\includegraphics[width=0.85\columnwidth]{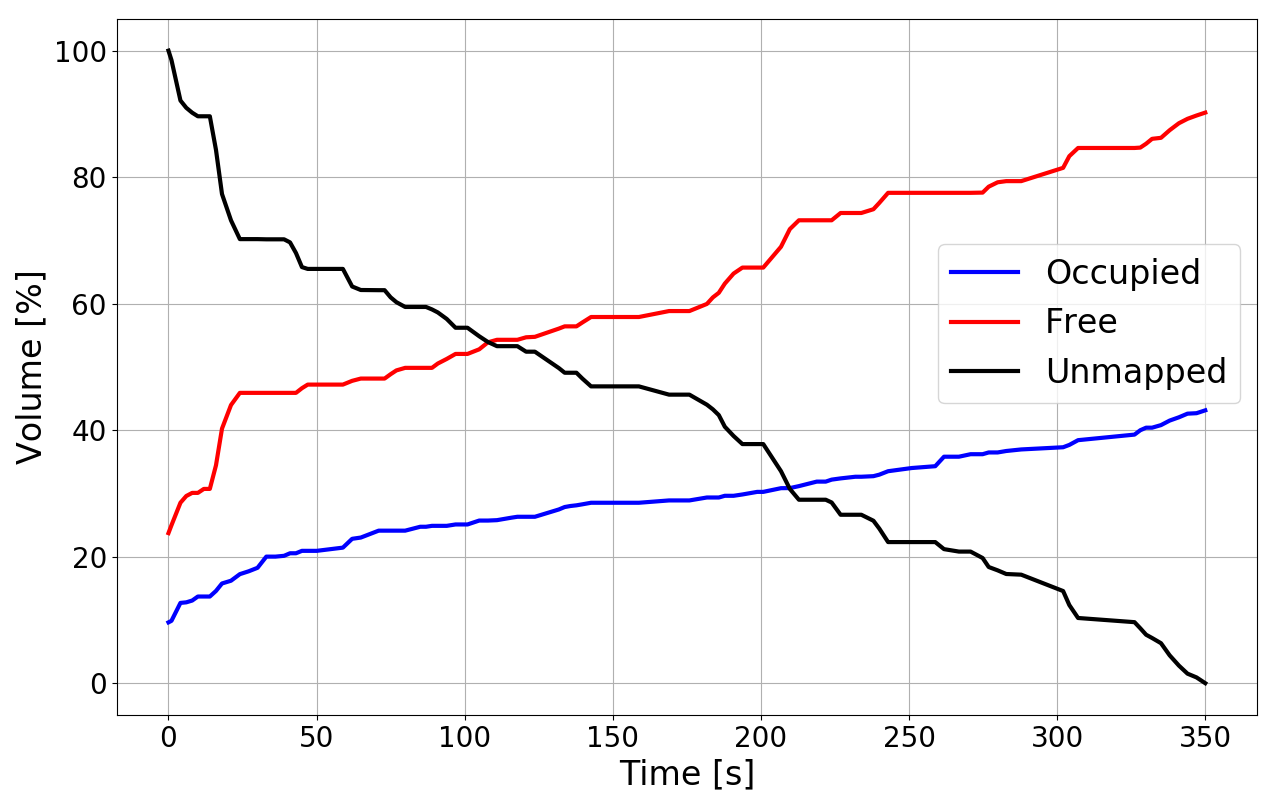}
	\caption{The  percentage  of  free,  occupied  and  unmapped volumes in time for the experimental scenario.}
	\label{fig: volume_borongaj}
\end{figure}

In the real world, we used the same parameters as in the large exploration scenario ($r_{max}$ = 0.5 m, $d_{exp}$ = 14 and the bandwidth = 2). Running the planner with the limited onboard resources and in real time, we were able to demonstrate fast exploration processing despite the large number of frontiers (Fig. \ref{fig:borongaj}). The OctoMap update time is much higher than in simulations because the rate of the sensor is higher, however the frontier detection time, which depends on $r_{max}$, is similar to times achieved in simulation. In the real world the average calculation time is 0.343s with a standard deviation of 0.043s. 
Fig. \ref{fig: volume_borongaj} shows that the total exploration time is about 350s. The result of the exploration is the OctoMap of the environment shown in  Fig. \ref{fig: path_borongaj}, in which the path traversed by the UAV during exploration is also shown.

Thorough comparison of experimental results with other state-of-the-art approaches is not possible, due to different environments, equipment and setup used. We briefly state, for completeness, experimental results available in the previously mentioned state-of-the art approaches. In \cite{Bircher2016} authors experiment in a 9x7x2m indoor arena with a MAV with $v_{max} = 0.25$m/s and a stereo camera and the exploration finishes after approximately after 250s. Size of our outdoor arena is 50x100x4m, and a UAV with $v_{max}=0.8$ m/s finishes the exploration in 350s. Regarding computation time, in \cite{Zhu2015} authors use RGBD camera in an office area of approximate size 10x10m, and obtain frontiers of size up to 200 cells. For these values, frontier detection takes about 18ms, clustering around 1ms and OctoMap update 0.5s. Frontier sizes in our experiments are 1-2 orders of magnitude larger, but the total average computation time is similar, 58ms. To showcase reproducibility of our results and facilitate more thorough future comparisons in the exploration field of research, data sets of simulations and experiments carried out for preparation of this paper are available\footnote{https://github.com/larics/exploration-datasets}. 

\section{Conclusion}
\label{sec:conclusion}

This paper deals with a novel multi-resolution frontier-based planner. The planner is capable of autonomously exploring a previously unknown area, creating an occupancy grid map using Cartographer SLAM and generating an OctoMap. Our approach may be applied to both indoor and outdoor environments. 

The results showed improvement in terms of total exploration and computation time suitable for real-time applications, despite the large input data sizes. A robust frontier detection speeds up the exploration process, while a novel clustering algorithm ensures target evaluation in the real time. 
This 3D exploration planner has been successfully tested in simulation scenarios, as well as in a real world experiment,  using  a  quadcopter equipped  with  a LiDAR. For future work we consider applying a more frequent waypoint assignment and a multi-robot system for exploration. 

Video recordings of frontier-based exploration for both simulation and experimental scenarios  can be found at YouTube \cite{video}.



\IEEEtriggeratref{30}
\bibliographystyle{ieeetr}


\begin{thebibliography}{10}

\bibitem{Yamauchi1997}
B.~Yamauchi, ``A frontier-based approach for autonomous exploration,'' in {\em
  Proceedings 1997 {IEEE} Int. Symposium CIRA97}, {IEEE} Comput. Soc. Press.

\bibitem{Hess2016}
W.~Hess, D.~Kohler, H.~Rapp, and D.~Andor, ``Real-time loop closure in 2d lidar
  slam,'' in {\em ICRA, 2016}, pp.~1271--1278, IEEE, 2016.

\bibitem{Zhu2015}
C.~Zhu, R.~Ding, M.~Lin, and Y.~Wu, ``A 3{D} frontier-based exploration tool
  for {MAVs},'' in {\em 27th Int. Conf. on Tools with Artificial Intelligence
  ({ICTAI})}, {IEEE}, November 2015.

\bibitem{Mannucci2017}
A.~Mannucci, S.~Nardi, and L.~Pallottino, ``Autonomous 3{D} exploration of
  large areas: A cooperative frontier-based approach,'' {\em MESAS},
  vol.~10756, pp.~18--39, Springer 2017.

\bibitem{Baiming2018}
T.~Baiming, S.~Jicheng, D.~Chaofan, and L.~Qingbao, ``A target point based
  {MAV} 3d exploration method,'' in {\em 2018 {IEEE} International Conference
  on Mechatronics and Automation ({ICMA})}, {IEEE}, August 2018.

\bibitem{Bircher2016}
A.~Bircher, M.~Kamel, K.~Alexis, H.~Oleynikova, and R.~Siegwart, ``Receding
  horizon "next-best-view" planner for 3{D} exploration,'' in {\em ({ICRA})},
  {IEEE}, May 2016.

\bibitem{Hornung2013}
A.~Hornung, K.~M. Wurm, M.~Bennewitz, C.~Stachniss, and W.~Burgard,
  ``{OctoMap}: an efficient probabilistic 3{D} mapping framework based on
  octrees,'' {\em Autonomous Robots}, vol.~34, pp.~189--206, Feb. 2013.

\bibitem{Comaniciu2002}
D.~{Comaniciu} and P.~{Meer}, ``Mean shift: a robust approach toward feature
  space analysis,'' {\em IEEE Transactions on Pattern Analysis and Machine
  Intelligence}, vol.~24, no.~5, pp.~603--619, 2002.

\bibitem{Alkhawaldah2012}
M.~Al-khawaldah and A.~N\"{u}chter, ``Multi-robot exploration and mapping with
  a rotating 3d scanner,'' {\em {IFAC} Proceedings}, vol.~45, no.~22,
  pp.~313--318, 2012.

\bibitem{Bachrach2009}
A.~Bachrach, R.~He, and N.~Roy, ``Autonomous flight in unknown indoor
  environments,'' {\em International Journal of Micro Air Vehicles}, vol.~1,
  pp.~217--228, December 2009.

\bibitem{GonzalezBanos2002}
H.~H. Gonz{\'{a}}lez-Ba{\~{n}}os and J.-C. Latombe, ``Navigation strategies for
  exploring indoor environments,'' {\em The International Journal of Robotics
  Research}, vol.~21, pp.~829--848, October 2002.

\bibitem{Joho2007}
D.~Joho, C.~Stachniss, P.~Pfaff, and W.~Burgard, ``Autonomous exploration for
  3d map learning,'' in {\em Autonome Mobile Systeme}, pp.~22--28, Springer,
  2007.

\bibitem{VasquezGomez2014}
J.~I. Vasquez-Gomez, L.~E. Sucar, R.~Murrieta-Cid, and E.~Lopez-Damian,
  ``Volumetric next-best-view planning for 3d object reconstruction with
  positioning error,'' {\em Int. Jour. of Adv. Robotic Systems}, vol.~11,
  p.~159, October 2014.

\bibitem{Dornhege2013}
C.~Dornhege and A.~Kleiner, ``A frontier-void-based approach for autonomous
  exploration in 3{D},'' {\em Advanced Robotics}, vol.~27, pp.~459--468, April
  2013.

\bibitem{Burgard2005}
W.~Burgard, M.~Moors, C.~Stachniss, and F.~Schneider, ``Coordinated multi-robot
  exploration,'' {\em {IEEE} Transactions on Robotics}, vol.~21, pp.~376--386,
  June 2005.

\bibitem{Orsulic2019}
J.~Orsulic, D.~Miklic, and Z.~Kovacic, ``Efficient dense frontier detection for
  2d graph {SLAM} based on occupancy grid submaps,'' {\em {IEEE} RAL},
  pp.~1--1, 2019.

\bibitem{Dornhege2011}
C.~Dornhege and A.~Kleiner, ``A frontier-void-based approach for autonomous
  exploration in 3d,'' in {\em 2011 {IEEE} International Symposium on Safety,
  Security, and Rescue Robotics}, {IEEE}, Nov. 2011.

\bibitem{Fukunaga1975}
K.~Fukunaga and L.~Hostetler, ``The estimation of the gradient of a density
  function, with applications in pattern recognition,'' {\em {IEEE}
  Transactions on Information Theory}, vol.~21, pp.~32--40, Jan. 1975.

\bibitem{Duda2001}
R.~O. Duda, P.~E. Hart, P.~E. Hart, and D.~G. Stork, {\em Pattern
  Classification}.
\newblock Wiley, 2001.

\bibitem{Arbanas2018}
B.~Arbanas, A.~Ivanovic, M.~Car, M.~Orsag, T.~Petrovic, and S.~Bogdan,
  ``Decentralized planning and control for {UAV}-{UGV} cooperative teams,''
  {\em Autonomous Robots}, vol.~42, pp.~1601--1618, feb 2018.

\bibitem{video}
``{A} {M}ulti-{R}esolution {F}rontier-{B}ased {P}lanner for {A}utonomous {3D}
  {E}xploration.''
  \url{https://www.youtube.com/playlist?list=PLC0C6uwoEQ8a88D6cKfa81Hfo_s_qVZxf}.

\end{thebibliography}

\end{document}